\documentclass[sigconf]{acmart}

\usepackage{booktabs} % For formal tables

\usepackage{caption}
\usepackage{subfig}
\usepackage{times}
\usepackage{graphicx}
\usepackage{amssymb,amsmath,amsfonts}

\usepackage{multirow}
\usepackage{comment}
\usepackage{setspace}
\usepackage{afterpage}
\usepackage{url}
\usepackage[noend]{algpseudocode}
\usepackage{fancyvrb}
\usepackage{tabularx}
\usepackage{enumitem}

\newcommand{\system}{Paranom}

\DefineShortVerb{\|}

\DefineVerbatimEnvironment%
  {Verb}{Verbatim}
  {framesep=2mm, rulecolor=\color{ForestGreen}}

\DefineVerbatimEnvironment%
  {VerbWithNumbers}{Verbatim}
  {fontsize=\small, framesep=2mm, rulecolor=\color{ForestGreen}, 
   numbers=none, numbersep=3pt, stepnumber=1}

\setcopyright{none}
\acmDOI{10.475/123_4}
\acmISBN{123-4567-24-567/08/06}
\acmConference[XXXXXX]{XXXXXXX}{XXXXXXX}{XXXXXXX}
\acmYear{2018}
\copyrightyear{2018}
\acmArticle{4}
\acmPrice{XXXX}

\begin{document}
\title{\system: A Parallel Anomaly Dataset Generator}
%\titlenote{Produces the permission block, and copyright information}
%\subtitle{Extended Abstract}
%\subtitlenote{The full version of the author's guide is available as \texttt{acmart.pdf} document}

\author{Justin Gottschlich}
\affiliation{%
  \institution{Intel Labs}
%  \city{Santa Clara}
%  \state{CA}
}
%\email{{justin.gottschlich, nesime.tatbul}@intel.com}

% The default list of authors is too long for headers.
\renewcommand{\shortauthors}{J. Gottschlich}

\begin{abstract}

In this paper, we present \system, a parallel anomaly dataset generator. We discuss its design and provide brief experimental results demonstrating its usefulness in improving the classification correctness of LSTM-AD, a state-of-the-art anomaly detection model.

\end{abstract}

% The code below should be generated by the tool at http://dl.acm.org/ccs.cfm. Please copy and paste the code instead of the example below.
%\begin{CCSXML}
% ...
%\end{CCSXML}
% \ccsdesc ...
%\keywords{Keywords, keywords}

\maketitle

\begin{abstract}

In this paper, we present \system, a parallel anomaly dataset generator. We discuss its design and provide brief experimental results demonstrating its usefulness in improving the classification correctness of LSTM-AD, a state-of-the-art anomaly detection model.

\end{abstract}

\vspace{-2mm}
\section{Introduction} \label{sec:introduction}

A \emph{dataset}, a collection of data usually manipulated as a single unit, is necessary for many machine learning (ML) techniques~\cite{russell:2003:ai}. In the context of deep learning, it has been shown that the larger and richer the dataset, the greater the potential accuracy of the model that can be built from it~\cite{goodfellow:2016:dl}. Because of this, possessing a large, high-quality dataset is usually a first step in building an ML model.

Contrarily, the practical development of ML models usually requires various sizes of data, beginning with a few items for initial model construction and up to billions of items for model deployment. In early development, small, yet rich, datasets can be useful because they enable rapid model tuning and topological changes without suffering significant performance penalties (e.g., days or weeks in training time). In practice, this means many ML models are generated iteratively by training and re-training on a growing dataset. If such manipulation is performed manually, it increases engineering overhead and the potential for data manipulation errors.

\emph{Anomaly detection}, the process of identifying outliers in a specific domain~\cite{aggarwal:2013:outlier, chandola:2009:adsurvey}, adds even more complexity to the dataset problem for at least two reasons. First, anomalies by definition are infrequent and therefore building an accurate anomaly detection model can be challenging due to the scarcity of anomalous data~\cite{chandola:2009:adsurvey}. Second, anomalies tend to be continuous events, which means data presented for them must usually be in a periodic, or time-series, ordered form~\cite{lavin:anomaly:2015, bailis:2017:macrobase}. For these reasons, and the growing importance of streaming systems, building large, rich, and meaningful datasets for anomaly detection is an open and increasingly important problem~\cite{singh-ijcnn17, bailis:2017:macrobase}.

In this paper, we present \emph{\system}, a \emph{par}allel \emph{anom}aly dataset generator. We make the following technical contributions:

\begin{enumerate}

\item A brief overview of \system's technical design.

\item An illustration of how \system's synthetic data can be used with LSTM-AD~\cite{malhotra:2015:esann}, a state-of-the-art anomaly detection model, improving its accuracy over using only real data. 

\end{enumerate}

\vspace{-2mm}
\section{Paranom's Design} \label{sec:design}

In this section we discuss \system's data uniqueness, data generation, data stochasticism, and parallel run-time execution model.

\begin{figure}[t]
\includegraphics[width=0.5\textwidth]{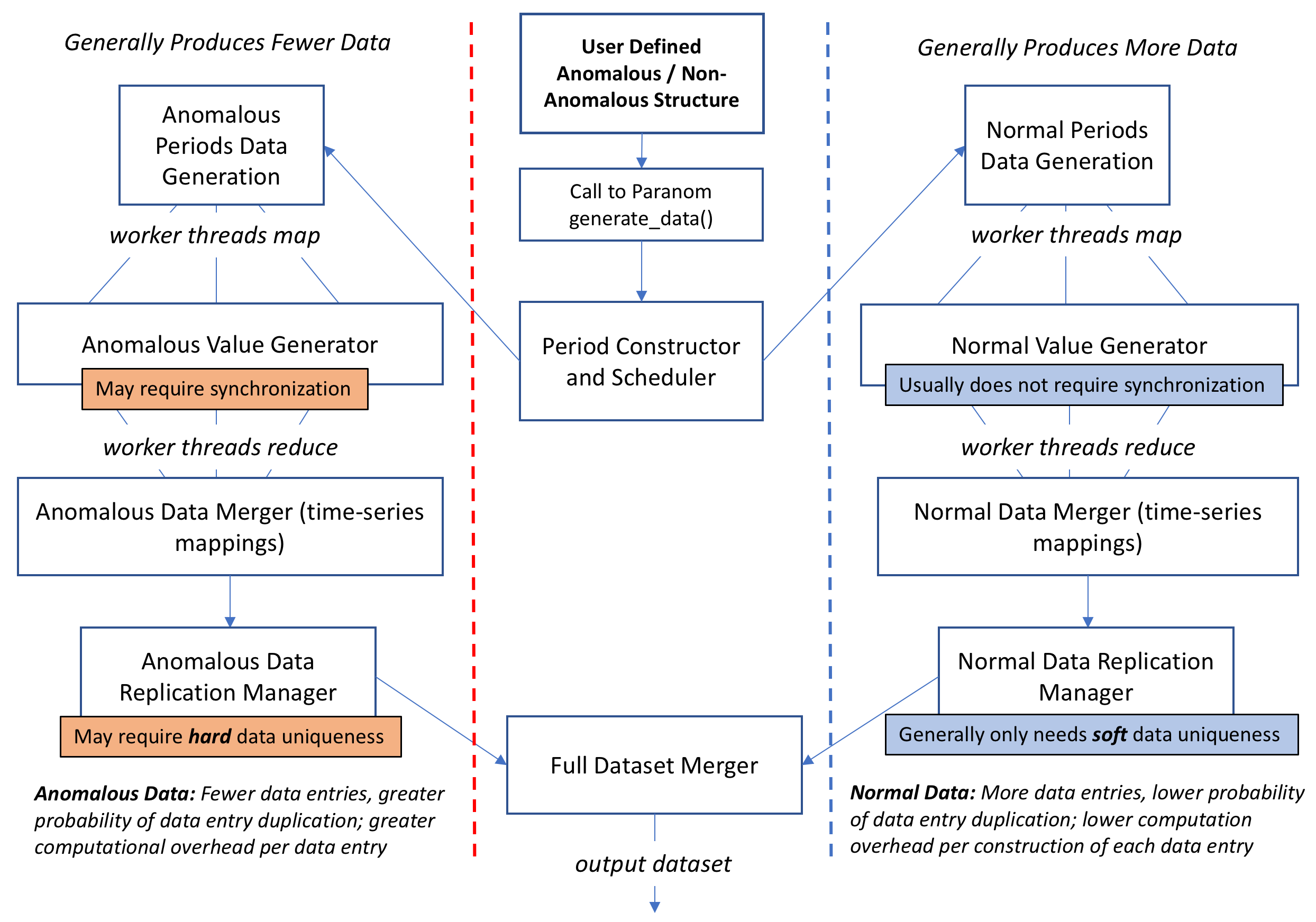}
\caption{\system's run-time execution model.}
\vspace{-2mm}
\label{fig:paranomDesign}
\end{figure}

\subsection{Data Uniqueness}

In synthetic data generation, data uniqueness cannot always be guaranteed. To illustrate this, consider an example where a user requests unique anomalous data, providing two possible discrete values: 0 and 1. Once both entries have been generated, it is impossible for any system to fulfill a third unique anomalous data point request.

To help manage this, \system\ provides two uniqueness controls for data generation:  \emph{hard} and \emph{soft}. We define hard uniqueness as a data property that must be met. If hard uniqueness is requested, and, if after a user-specified number of tries \system\ has unsuccessfully generated a unique datum, \system\ will terminate execution. We define soft uniqueness as a data property that might be met, but, if after a user-specified number of tries, it has not been met, \system\ will continue execution using its last generated datum entry.

Because anomalous data may be scarce, duplication of any of its already small number of data points may result in model over-fitting~\cite{bishop:2006:prml}. To address this, \system\ provides two data uniqueness controls: one for normal data and one for anomalous data. This enables a user to generate soft unique non-anomalous data and hard unique anomalous data, or vice-versa, as needed.

\subsection{Data Generation}

\system\ supports the following two ways of data generation.~\footnote{Due to limited space, we have omitted code examples.}

\noindent
{\bf Stochastic Variables.} These variables support controlled randomization, where developers define them with a specified range of stochasticism for both the anomalous and non-anomalous values they will generate. \system\ then handles all value generation.

\noindent
{\bf Callback Variables.} If a developer requires full control over value generation, she can define a callback variable, which requires the construction of two callback functions for each variable, one for anomalous data generation and one for non-anomalous. At run-time, the appropriate callback function will be invoked for value generation for each variable at each unique discrete timestamp.

\subsection{Data Stochasticism}
\system\ provides seedable randomization, which (if needed) can ensure repeatable stochasticism in dataset generation. This can be useful for iterative ML training from small to large datasets, where the previously seen data is guaranteed to remain unchanged as the dataset grows in size. \system\ also provides controls for the anomalous and non-anomalous data that will be randomly present in a dataset. In addition to providing controls for the absolute number of data points, \system\ also provides controls for the stochastic frequency of anomalous and non-anomalous data.

\subsection{Parallel Run-Time Execution}

To ameliorate the performance overhead of possibly generating billions of entries for a dataset, \system\ was designed with the goal of being \emph{perfectly parallel}, where there is no multithreaded synchronization used in the generation of its data (see Figure~\ref{fig:paranomDesign})~\cite{herlihy:2008:amp}. Although \system's data generation is perfectly parallel, user level synchronization can be used, if needed, within the user-defined callback variables. This can be useful in generating unique data and sequentially dependent time-series data, among other things.

\begin{figure}[t]
\includegraphics[width=0.45\textwidth]{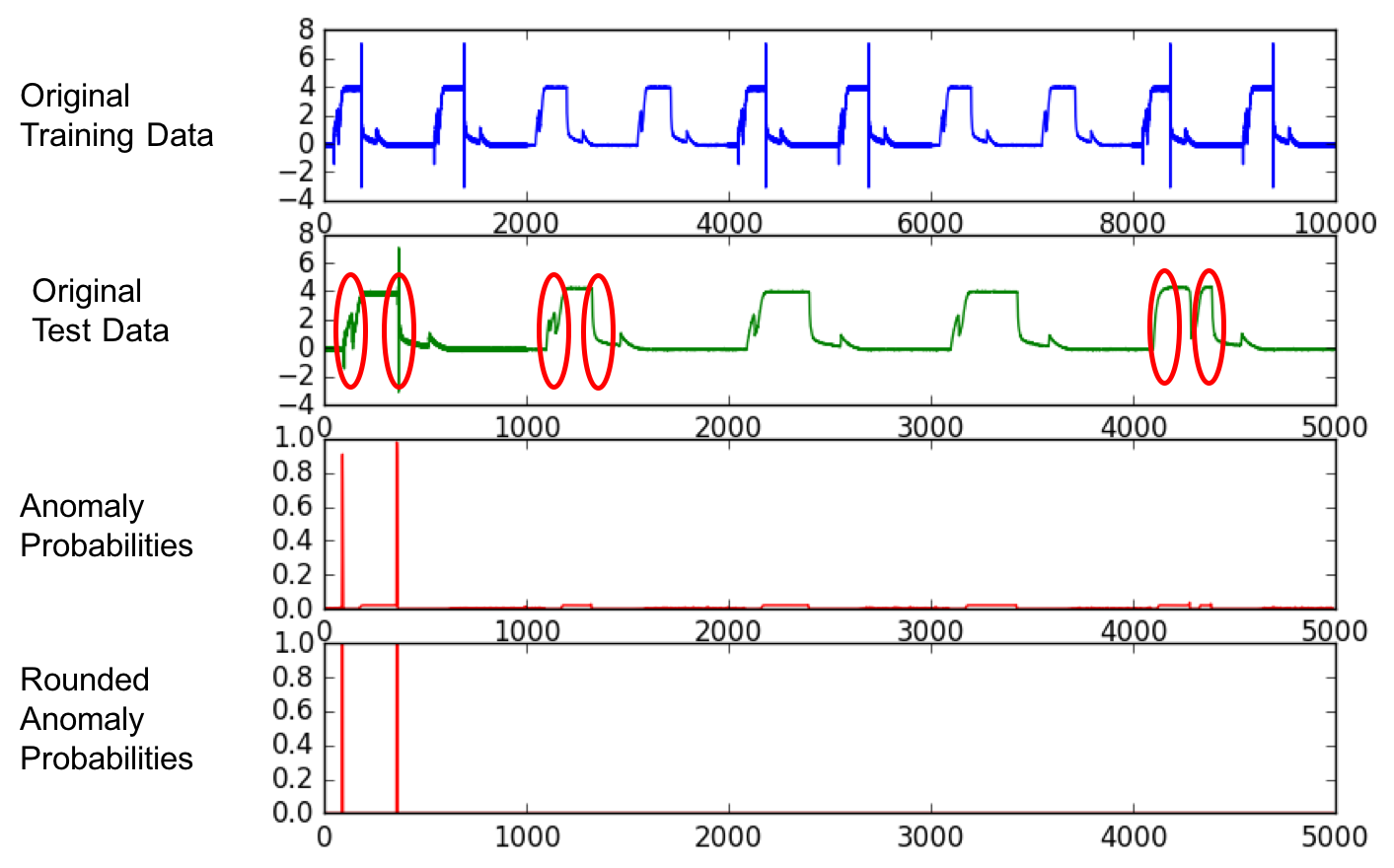}
\caption{Visualization of original LSTM-AD space shuttle training, testing and prediction data (red ovals are anomalies).}
\label{fig:originalVisual}
\end{figure}

\section{Experimental Evaluation} \label{sec:experiments}

In this section, we describe how we used \system\ to improve the accuracy of LSTM-AD, a state-of-the-art anomaly detection ML model by Malhotra et al., compared to using only real data for a space shuttle valve sensor anomaly~\cite{malhotra:2015:esann}.~\footnote{We have used \system\ to generate data for over a dozen different anomalous scenarios. However, due to limited space, we include only one example from Malhotra et al.'s LSTM-AD paper~\cite{malhotra:2015:esann}. \system\ can also be used to generate only non-anomalous data which can be useful for zero-positive learning~\cite{mejbah:2017:autoperf}.}

\noindent
{\bf Experimental Setup.} 
We trained two LSTM-AD models in TensorFlow~\cite{tensorflow-osdi16}. One used only real space shuttle data. The other used portions of the real non-anomalous data in conjunction with anomalous data solely generated by \system as described below.

\noindent
{\bf Similarities.} Both models were constrained to the same training data size (i.e., 10,000 data points) and tested against the same real test data. All aspects of the LSTM-AD model (e.g., topology, activations, etc.), as well as the training iterations, were identical in both settings.

\noindent
{\bf Differences.} The model trained using \system's generated training data did not use any of the real-world anomalous training data. Instead, we created a \system\ callback variable that would generate anomalous data uniquely different from the real-world non-anomalous data with a stochastically chosen value range. We then had \system\ inject synthetic anomalies with a $1\%$ frequency and a variable anomaly duration range similar to its real anomalous events. 

\begin{figure}[t]
\includegraphics[width=0.45\textwidth]{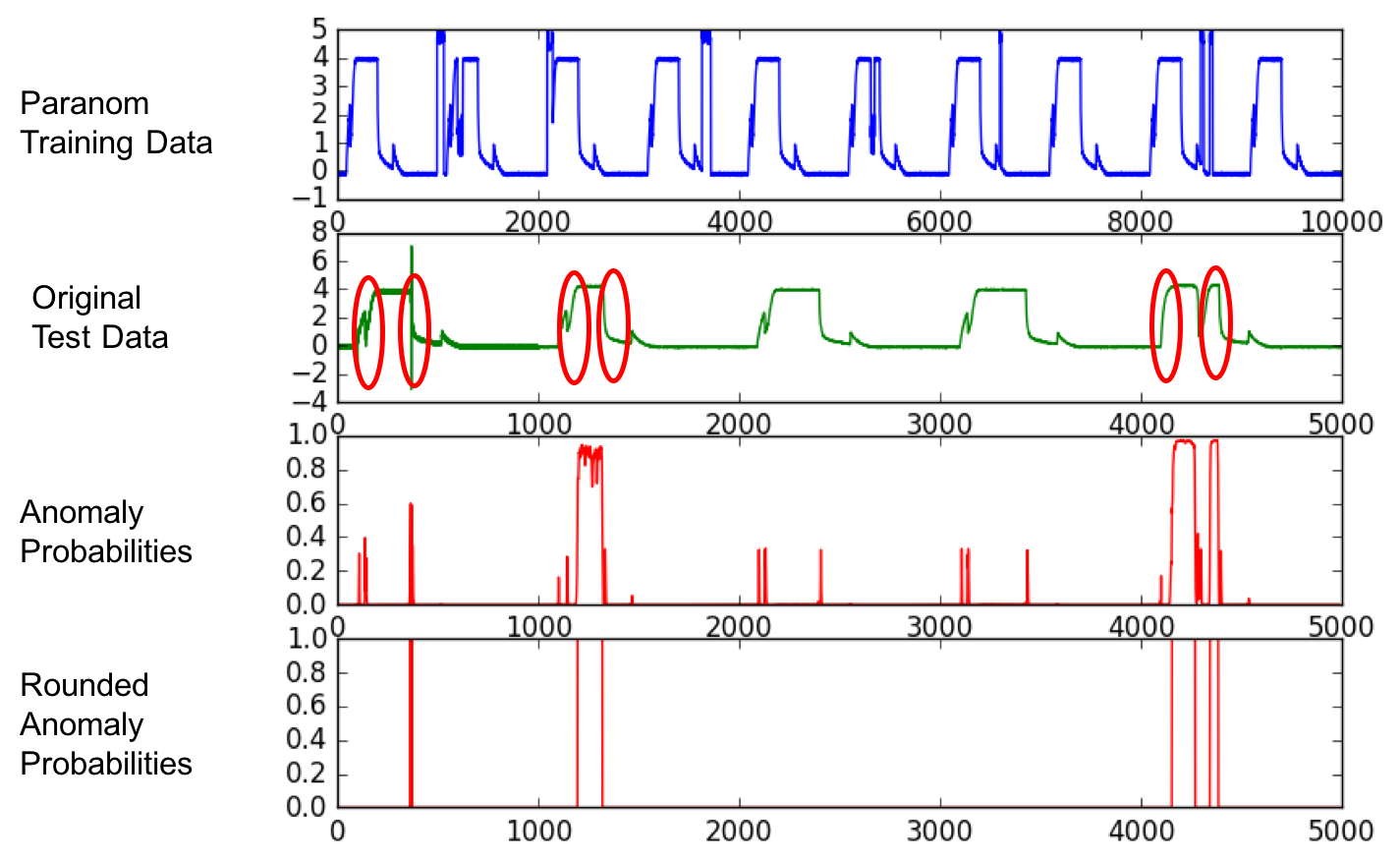}
\caption{Visualization of Paranom LSTM-AD space shuttle training, testing and prediction data (red ovals are anomalies).}
\label{fig:paranomVisual}
\end{figure}

\noindent
{\bf Results.} 
The differences between the original training data and our \system\ generated data can be seen in Figures~\ref{fig:originalVisual} and \ref{fig:paranomVisual}, respectively. Figure~\ref{fig:originalVisual} shows the original data used to train LSTM-AD, including six real anomalies (denoted by red ovals). It also shows the LSTM-AD predictions against the test data after being trained against the original data. Figure~\ref{fig:paranomVisual} shows our \system\ generated training data, which includes \system's synthetically generated stochastic anomalies. Once trained against the real training data and \system's training data, we tested both LSTM-AD models against the original testing data. The original model identified two of the six anomalies. The \system\ model identified five of the six anomalies.

\begin{figure}[h!]
\includegraphics[width=0.49\textwidth]{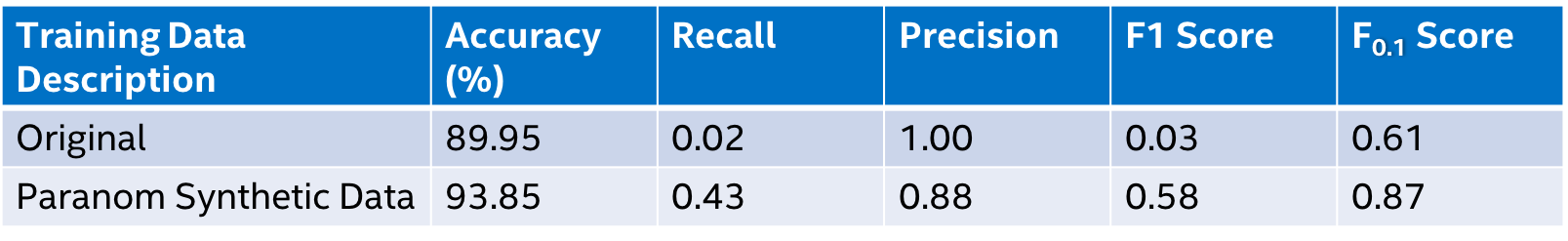}
\caption{The performance results of LSTM-AD when using its original training data versus using \system's training data.}
\label{fig:perf}
\end{figure}

As can be seen in Figure~\ref{fig:perf}, the \system\ LSTM-AD model had improved accuracy, recall, $F_1$, and $F_{0.1}$ score when compared to the model trained against only real data. The only LSTM-AD result that was not improved was precision. This is because the \system\ LSTM-AD model introduced some false positives.

\section{Conclusion}  \label{sec:conclusion}

In this paper, we briefly presented \system's design and parallel execution model. We provided an empirical illustration showing the benefit of using \system's synthetically created data to improve the robustness of LSTM-AD, a state-of-the-art anomaly detection ML model, by an order of magnitude for recall and \emph{$F_1$} using \system's data over using only real data. 

%\system's performance improvements to LSTM-AD required no model changes nor data expertise in the particular problem domain.

\bibliographystyle{ACM-Reference-Format}
\bibliography{anomaly}

\end{document}